\documentclass[sigconf]{acmart}

\def\BibTeX{{\rm B\kern-.05em{\sc i\kern-.025em b}\kern-.08emT\kern-.1667em\lower.7ex\hbox{E}\kern-.125emX}}
 
 \usepackage{balance}
  
\usepackage{xcolor}
\newcommand{\bm}[1]{\mathbf{#1}}

\copyrightyear{2019} 
\acmYear{2019} 
\acmConference[MM '19]{Proceedings of the 27th ACM International Conference on Multimedia}{October 21--25, 2019}{Nice, France}
\acmPrice{15.00}
\acmDOI{10.1145/3343031.3350854}
\acmISBN{978-1-4503-6889-6/19/10}

\begin{document}
	\fancyhead{}
	
	\title[LinesToFacePhoto with CSAGAN]{LinesToFacePhoto: Face Photo Generation from Lines with Conditional Self-Attention Generative Adversarial Network}

	 \author{Yuhang Li}
	 \email{lyh9001@mail.ustc.edu.cn}
	 \affiliation{%
     \institution{NEL-BITA, University of Science and Technology of China}
	 }
 
	 \author{Xuejin Chen}
	 \authornote{Xuejin Chen is the corresponding author.}
	 \email{xjchen99@ustc.edu.cn}
	 \affiliation{%
	 	\institution{NEL-BITA, University of Science and Technology of China}
 	}
 	 	\author{Feng Wu} 
 	 	\email{fengwu@ustc.edu.cn}
 	 	\affiliation{%
 	 		\institution{NEL-BITA, University of Science and Technology of China}
 	 	}
 	 	 
 	 \author{Zheng-Jun Zha} 
 	 \email{zhazj@ustc.edu.cn}
 	 \affiliation{%
 	 	\institution{NEL-BITA, University of Science and Technology of China}
 	 }
	\renewcommand{\shortauthors}{Y. Li et al.}
	
	\begin{abstract}
		In this paper, we explore the task of generating photo-realistic face images from lines.
		Previous methods based on conditional generative adversarial networks (cGANs) have shown their power to generate visually plausible images when a conditional image and an output image share well-aligned structures. 
		However, these models fail to synthesize face images with a whole set of well-defined structures, e.g. eyes, noses, mouths, etc., especially when the conditional line map lacks one or several parts. 
		To address this problem, we propose a conditional self-attention generative adversarial network (CSAGAN).
		We introduce a conditional self-attention mechanism to cGANs to capture long-range dependencies between different regions in faces.
		We also build a multi-scale discriminator. The large-scale discriminator enforces the completeness of global structures and the small-scale discriminator encourages fine details, thereby enhancing the realism of generated face images. 
		We evaluate the proposed model on the CelebA-HD dataset by two perceptual user studies and three quantitative metrics.
		The experiment results demonstrate that our method generates high-quality facial images while preserving facial structures. Our results outperform state-of-the-art methods both quantitatively and qualitatively.
	\end{abstract}

	
	%
	\begin{teaserfigure}
		\includegraphics[width=\textwidth]{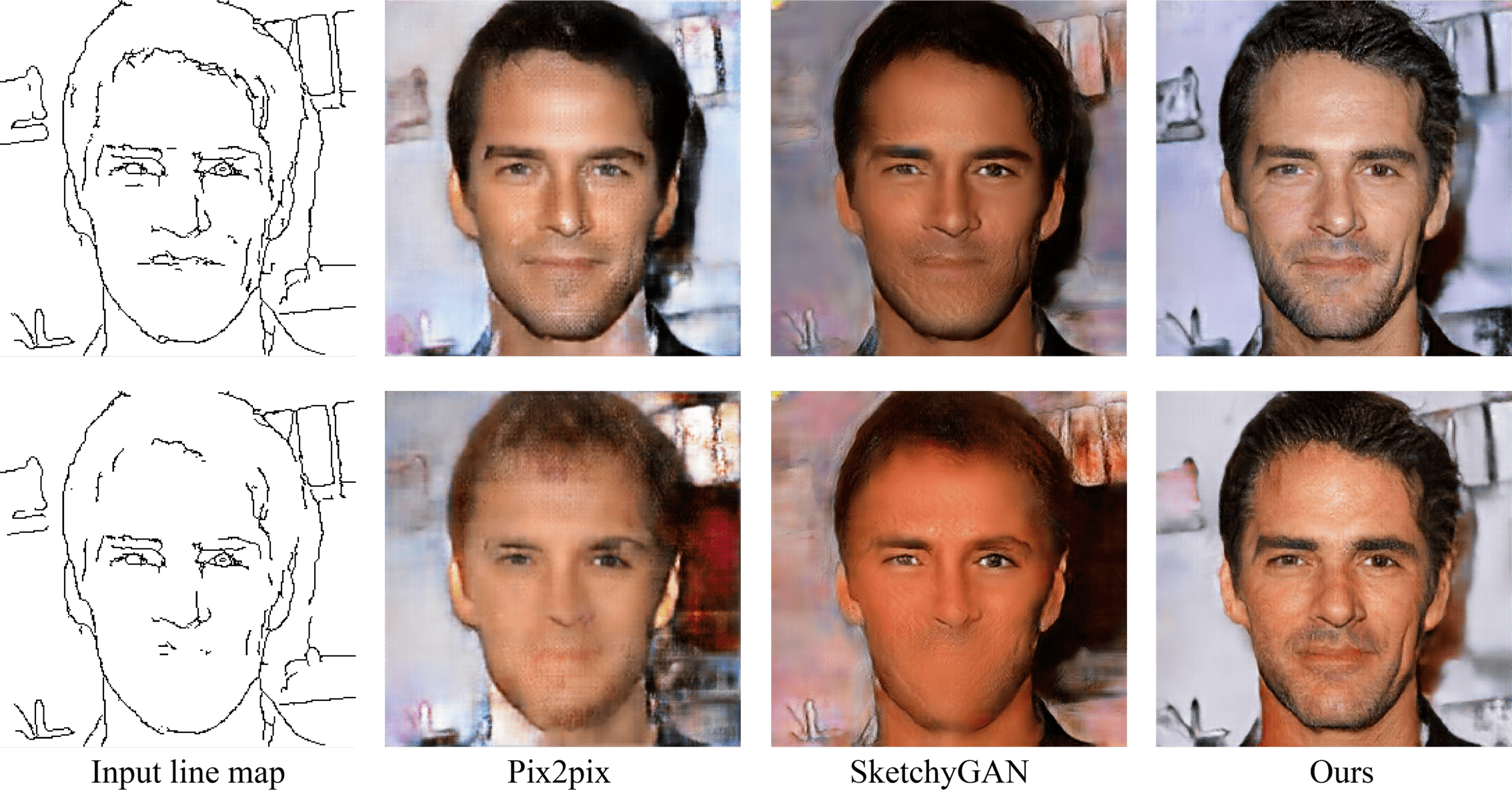}
		\caption{From sparse lines that coarsely describe a face, photorealistic images can be generated using our conditional self-attention generative adversarial network (CSAGAN). With different levels of details in the conditional line maps, CSAGAN generates realistic face images that preserve the entire facial structure. Previous works~\cite{pix2pix, SketchyGANs} fail to synthesize certain structural parts (i.e. the mouth in this case) when the conditional line maps lack corresponding shape details.}  
		\label{fig:teaser}
	\end{teaserfigure}
	
	%
	\maketitle
	
	\section{Introduction}

When creating something from scratch, a natural and intuitive way is to draw lines. 
Line drawing is an effective form of visual thought.
It describes the structure and shape of a desired object more specifically than text. 
Turning lines into photorealistic images has drawn a lot of attention in computer graphics and computer vision for many years. 
Benefiting from the massive amounts of images on the Internet, many approaches in the past decade have been proposed based on line-based image retrieval and image synthesis techniques~\cite{Sketch2Photo,Photosketcher,Xu-Sketch2Scene2013,Turmukhambetov:2015:SketchImageSynthesis}.
While these methods successfully maintain the primary structure of the desired object or scene, they typically fail to generate fine details due to the limited capability of traditional image synthesis techniques.

With the emergence of deep neural networks (DNN), a series of approaches based on generative adversarial networks (GANs) have been proposed for realistic image synthesis~\cite{VAEs, PixelCNN}.  
A generator and a discriminator in GANs are trained by playing a min-max game to guide the generated samples to become indistinguishable from real ones. 
Image-to-image translation, which is a specific application of conditional GANs, aims to translate an image in one domain to a target image in another domain, while preserving main contents and structures in these two images. 
Since the first image-to-image translation model, pix2pix~\cite{pix2pix}, was proposed, there have been many variants in both supervised and unsupervised manners~\cite{CycleGANs, DualGANs,CoupleGANs,BicycleGANs,pix2pixHD,PGGAN}. 
These models successfully synthesize realistic textures when complete and detailed structures are given in the conditional image.

However, when the structure is partially provided in the conditional image, which is exactly the case of line drawings or edge maps, previous models fail to complete the missing structure.
This is mainly because these methods strictly follow the provided edges when synthesizing the generated image; thus, they do not generate new structures at the place where few edges are provided. 
Since faces are composed of well-defined structural parts, e.g. noses, mouths, eyes, etc., the synthesized face images should contain the whole set of these structural parts to appear realistic, even when the conditional line maps lack edges around the supposed locations of these parts. 
As Figure~\ref{fig:teaser} shows, using our method, the generated images from two line maps with different levels of details are more realistic because of the complete global structures and fine textures. 
Previous methods~\cite{pix2pix, SketchyGANs} fail to render realistic face images while edges around the mouth area are incomplete.

The underlying reasons for this failure are mainly two-fold.
First, existing GANs are built primarily on convolutional layers. Since the convolutional operator has a local receptive field depending on its kernel size, a large receptive field is achieved by stacking multiple convolutional layers. However, it is non-trivial for current network optimizers to discover proper parameter values that model the long-range dependence through several convolutional layers~\cite{SAGANs}.  
Secondly, existing discriminators used in GANs focus on examining local patches instead of capturing the global information; therefore, they fail to enforce the generator to synthesize global structure of the generated image. 

Considering the first issue, we propose a conditional self-attention mechanism to the image-to-image translation model generator to address the problem.
Self-attention, which computes the response at one position as a weighted sum of the features at all positions, is able to capture long-range dependency across different parts~\cite{Non-local, Attention, MachineReading, SAGANs}.  
In order to adapt the conditional setting of image-to-image translation and encourage the GAN model to fully exploit the information from the conditional image directly, we propose a \emph{conditional self-attention module (CSAM)}, which enables the higher layers to sense information from the conditional image and capture long-range dependency. 
For the second issue, we establish a multi-scale discriminator to capture information from different levels. The small-scale discriminator has a local receptive field and improves the fine textures of local patches, while the large-scale discriminator ensures the completeness of the global structure in the generated images.

In this paper, we focus on the task of portrait photograph generation from line drawings,  while preserving well-defined face structures, which are critically important for the realism of face photos. 
Our contributions are summarized as follows:
\begin{enumerate}
\item We first introduce the self-attention mechanism to line-to-image translation and propose a novel conditional self-attention generative adversarial networks. Unlike convolution-based methods, the proposed model is able to model long-range dependence and global structures in face images.
 
\item We show the effectiveness of the proposed model by a series of experiments on the CelebA-HD dataset. Our method generates high-quality face images from sparse lines and preserves facial structures. The proposed CSAGAN outperforms state-of-the-art methods both quantitatively and qualitatively.

\end{enumerate}

	\section{Related work} 
\label{sec:related_work}
Our CSAGAN method for generating face photos from line maps is built upon previous work on image-to-image translation, attention mechanism, and line-based image synthesis.
We discuss most related techniques in this section.
\begin{figure*}
	\includegraphics[width=\textwidth]{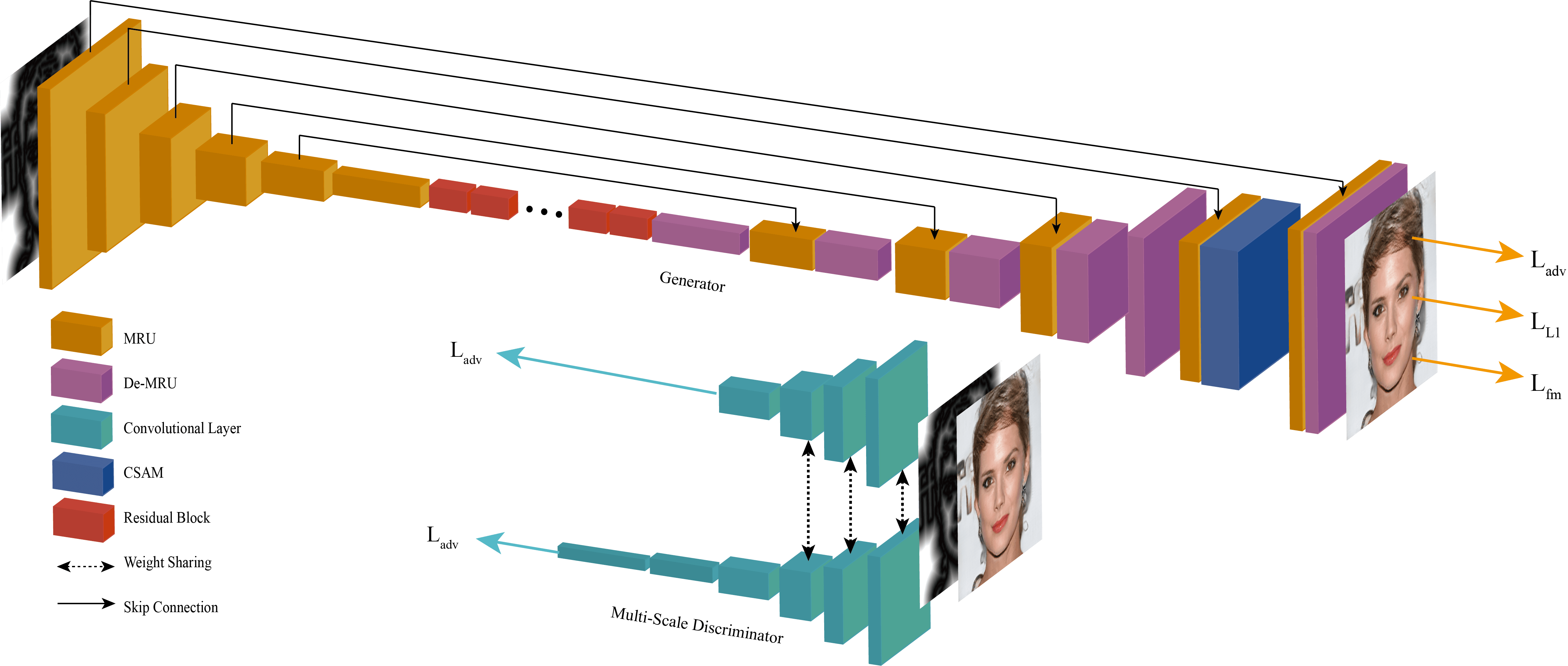}
	\caption{The architecture of our model. The proposed CSAM (the blue block) is added before the last convolutional layer. The multi-scale discriminator (only two are drawn) are applied to encourage the generator to produce realistic results with complete structure and delicate textures.}
	\label{fig:architecture}
\end{figure*}

\subsection{Image-to-Image Translation with GANs}
Given an image in one domain, image-to-image translation methods generate a corresponding image in another domain, while depicting the same scene or object in different styles.
The pix2pix method~\cite{pix2pix} first introduced image-to-image translation with conditional GAN for a range of applications. 
However, it is difficult for the convolution-based architecture used in pix2pix to discover long-range dependence across different regions.
Moreover, the patch-wise discriminator in pix2pix can not ensure that global structures are well captured.
Following pix2pix, many supervised techniques have been proposed to improve the resolution and details of target images.
\cite{outdoor_scene} studied how to generate images of outdoor scenes from semantic label maps coupled with attributes. 
\cite{BicycleGANs} presented a framework that encourages the connection between the output and the latent code to be invertible so as to model the multi-modal distribution of possible outputs.
\cite{CRN, pix2pixHD} used coarse-to-fine refinement frameworks to synthesize high-resolution photographic images from semantic label maps. 
In comparison, our work focuses on translating rough lines to realistic face photos, in which the key challenge is to learn the global structure and long-range dependence across different regions in a face image. 

\subsection{Self-Attention Mechanism}
In order to sense global structures in a large receptive field, several convolutional layers and large kernel sizes have typically been required in previous GAN-based techniques. 
However, simply stacking convolutional layers or increasing kernel sizes seriously harms the computational efficiency. 
Self-attention, which computes the response at one position as a weighted sum of the features at all positions, is able to capture long-range dependence across different parts. 
\cite{Attention} applied self-attention to capture global dependence in sequential data for machine translation, and they demonstrated the plausible effectiveness of the self-attention mechanism. 
\cite{Transformer} studied on combining the self-attention mechanism with autoregressive models, and proposed an image transformer model for image generation. 
Inspired by non-local operations in computer vision, \cite{Non-local} utilized the self-attention mechanism as a non-local operation to model long-range spatial-temporal dependence for video processing.
\cite{SAGANs} introduced self-attention to unconditional GANs and showed its advantages in generating natural images from noise vectors. 
\cite{SARN} focused on saliency detection, utilizing a recurrent structure for shallow layers and a self-attention module for deep layers. 
\cite{RAM} proposed a novel parallax version of self-attention for stereo image super-resolution. 
Inspired by previous works, we first explore the self-attention mechanism in the context of lines-to-photo translation to exploit global structures and long-range dependence between different parts in faces. 

\subsection{Lines-based Synthesis}

Synthesizing images and models~\cite{SketchingReality} from strokes or lines is not a novel idea.
Earlier techniques of image synthesis from lines~\cite{Photosketcher, Sketch2Photo} search for image patches in a large-scale database using the drawn lines, and then they fuse the retrieved image patches. 
With recently developed GANs, image-to-image translation techniques have been applied to the edge-to-photo task~\cite{pix2pix,pix2pixHD}.
However, these general frameworks, which are not specially designed for line drawings, require input edge maps that contain complete and carefully drawn lines to generate visually pleasing results. 
Taking hand-drawn sketches as input, SketchyGAN synthesizes plausible images for 50 object categories~\cite{SketchyGANs} . 
A masked residual unit (MRU) is proposed to improve the information flow by injecting the input image at multiple scales.
However, when the conditional line maps lack specific structural
parts, these GAN-based methods suffer from incomplete structures in the generated
images.
In comparison, our method learns long-range dependence in face images and produces photo-realistic images from line maps of different detail levels.

\section{Method}
\label{sec:method}
%
%
In this section, we introduce our Conditional Self-attention Generative Adversarial Network (CSAGAN) for translating line maps to photo-realistic photos of human faces.
The architecture of our model is shown in Figure~\ref{fig:architecture}. 
The generator is based on an encoder-decoder architecture with residual blocks. Skip connections~\cite{Unet} are applied between corresponding layers in the encoder and decoder. 
We adopt masked residual units (MRUs)~\cite{SketchyGANs} in our framework. 
We add our proposed conditional self-attention module (Sec.\ref{subsec:CSAM}) before the last MRU to model long-range dependence among feature maps. 
The conditional line map is resized and concatenated into feature maps at multiple scales to use as input for the MRUs and CSAM. 
Finally, to encourage the generator to produce realistic face images with complete structures and fine textures, we use a multi-scale discriminator (Sec.~\ref{subsec:disciminator}) to classify a face image globally and image patches locally as real or synthesized.

\begin{figure}
	\includegraphics[width=\linewidth]{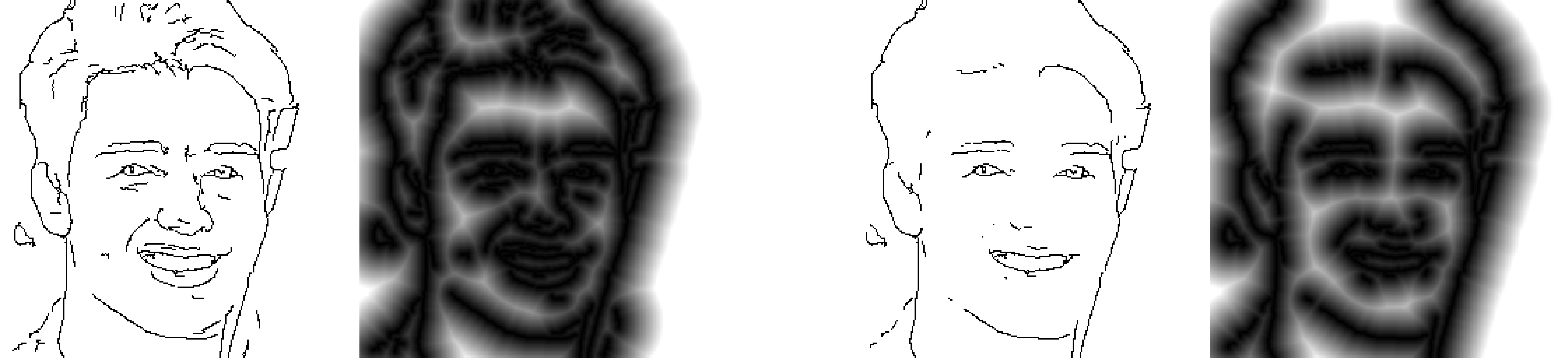}
	\caption{Dense distance field representation of sparse line maps.}
	\label{fig:DF}
\end{figure}

\paragraph{Input to CSAGAN} 
Since line maps are very sparse and rough, we adopt a dense representation using a distance transform.
From a black-white line map, an unsigned Euclidean distance field is calculated as the conditional image. 
Figure~\ref{fig:DF} shows two examples of the distance fields generated from two line maps with different levels of details. Compared to the sparse and rough line maps, the dense distance fields spread the shape information to all pixels so that the extracted feature maps are more robust to incompleteness and noise in the input line maps.
Similar ideas of using distance field representations can be found in several sketch-based applications~\cite{repair_3d, shape_completion, SketchyGANs}.
On the other hand, some conditional GANs add a noise vector to the generator as input to avoid producing a deterministic output. 
However, the pix2pix model has shown that the noise vector is ignored by the generator and hardly changes the output. 
We observe the same phenomenon in our experiments, thus we do not apply noise vectors in our model.

%

%
\subsection{Conditional Self-Attention Module (CSAM)}
\label{subsec:CSAM}
Inspired by SAGANs~\cite{SAGANs}, we propose a conditional self-attention module (CSAM) for our lines-to-photo translation task to extract long-range dependence. 
This module is designed as a general module of conditional frameworks and can be added after any existing conditional modules of feature extraction. 
Given feature maps extracted from the previous layer $\bm{a}\in \mathbb{R}^{C\times H\times W}$ and a resized conditional line map $\bm{x}\in \mathbb{R}^{1\times H\times W}$ that matches the resolution of the current layer of feature maps.
we concatenate them to get $[\bm{a}, \bm{x}]$ as conditioned features, where $[\cdot,\cdot]$ is the concatenation operation, and $C$, $H$, and $W$ are the number of channels, height, and width of the feature map $\bm{a}$. 
This allows the network to form the attention based on the conditional image as well as the feature maps.
In order to calculate the attention, we map the conditional features $[\bm{a}, \bm{x}]$ to two feature spaces:
\begin{equation}
	\label{eqn:f}
	f([\bm{a}, \bm{x} ])=\mathbf{W}_f[\bm{a}, \bm{x} ],
\end{equation}
\begin{equation}
	\label{eqn:g}
	g([\bm{a}, \bm{x} ])=\mathbf{W}_g[\bm{a}, \bm{x} ],
\end{equation}
where $\mathbf{W}_f$, $\mathbf{W}_g\in \mathbb{R}^{\hat{C}\times (C+1)}$ are trainable weights and are implemented by $1\times 1$ convolutions.  
Here, we use $\hat{C}=C/8$ in our experiments following the setting of SAGAN \cite{SAGANs}.

Let $\mathbf{B}\in \mathbb{R}^{N\times N}$ be the attention map, where $N=H\times W$. Every element in $\mathbf{B}$, denoted as $b_{j,i}$, indicates the extent to which the model attends to the $i^{th}$ pixel while synthesizing the $j^{th}$ pixel. $b_{j,i}$ is calculated by 
\begin{equation}
	\label{eqn:beta}
	b_{j,i}=\frac{exp(s_{ij})}{\sum^{N}_{i=1}exp(s_{ij})},
\end{equation}
in which $s_{ij}=f([\bm{a}, \bm{x} ])^Tg([\bm{a}, \bm{x} ])$. 
Next, we use $b_{j,i}$ as the attention weights and compute the response map $\bm{r}=(\bm{r}_1, \bm{r}_2,\cdots, \bm{r}_{N})\in \mathbb{R}^{C\times N}$ at every position as a weighted sum of the features at all positions, where
\begin{equation}
	\label{eqn:response}
	\bm{r}_j=\sum^{N}_{i=1}b_{j,i}h([\bm{a}, \bm{x}]),
\end{equation}
where $h([\bm{a}, \bm{x}])=\mathbf{W}_h[\bm{a}, \bm{x} ]$ and $\mathbf{W}_h\in \mathbb{R}^{C\times (C+1)}$.
As suggested in \cite{SAGANs}, we further multiply the response of the attention layer by a scale parameter $\gamma$ and add it back to the input feature maps. The final output is calculated by 
\begin{equation}
	\label{eqn:output}
	\bm{o}_i=\gamma \bm{r}_i+\bm{a}_i,
\end{equation}
where $\gamma$ is a trainable value and is set to $0$ at the beginning of the training process. 
In this way, the network learns local dependence at early stages in the training process, and then it learns long-range dependence by assigning more weights to the non-local evidences progressively.
\begin{figure}
	\includegraphics[width=0.7\linewidth]{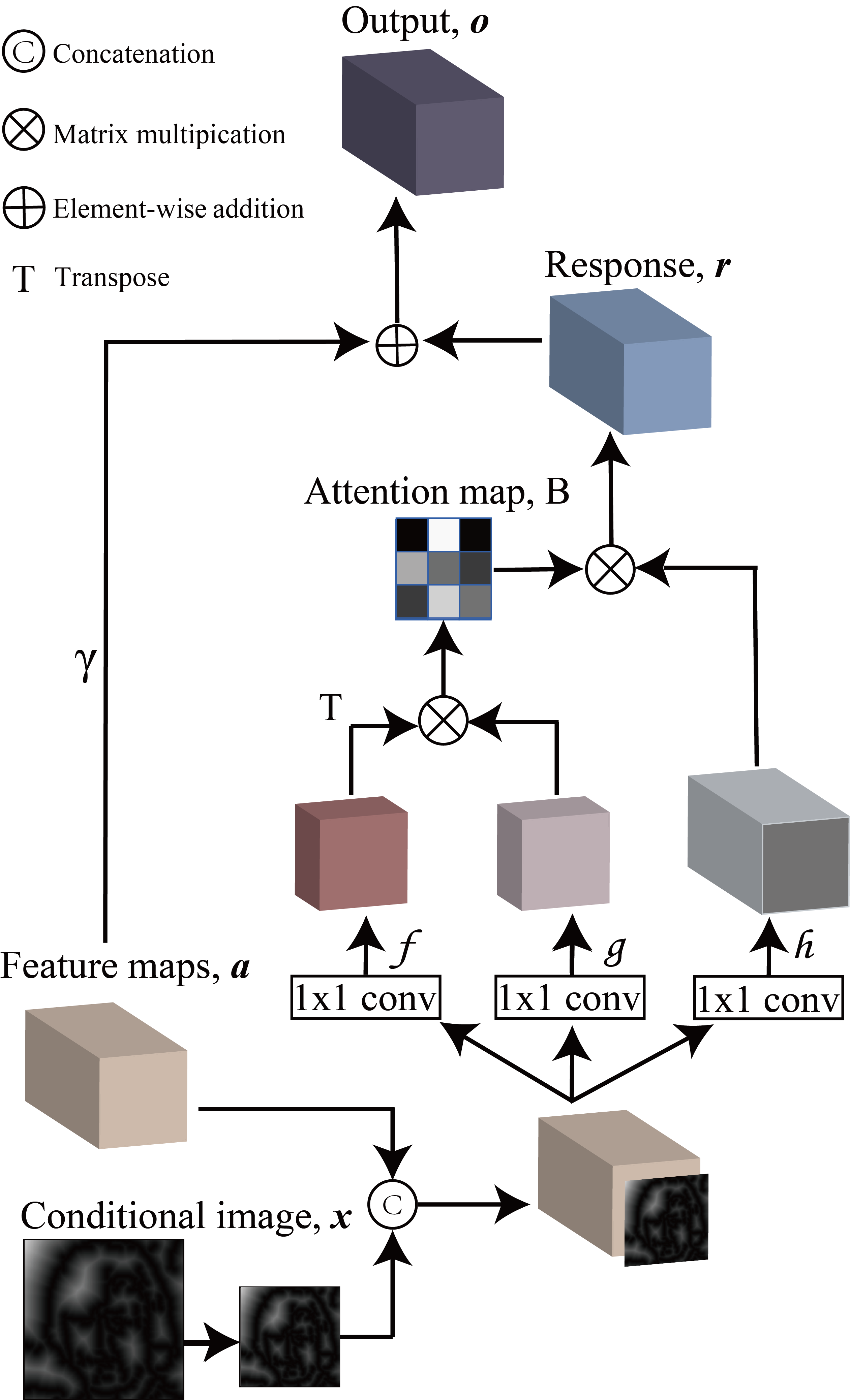}
	\caption{Components of the proposed CSAM. Given the conditional image and feature maps from the previous layer, the output feature maps are calculated in a self-attention manner. }
	\label{fig:CSAM}
\end{figure}
%

%
\subsection{Multi-Scale Discriminator}
\label{subsec:disciminator}
The discriminators for the pix2pix model and SketchyGAN are patch-wised, which distinguish real/fake images patch by patch convolutionally in local receptive fields that are much smaller than the size of the input images. 
The average value of all responses is calculated as the ultimate output of the patch discriminator.
This is based on the assumption of independence between pixels separated by more than one patch's diameter. 
However, since the structural constraint is global information across an entire image, the patch-wise discriminator fails to capture the global structure.
We design a multi-scale discriminator consisted of $N_D$ subnetworks with different depths and, therefore, different sizes of receptive fields in their last layers. The receptive field in the last layer of the deepest subnetwork is as large as the entire image to capture the global structure. These subnetworks share weights with each other in first few layers since the lower-level features of these discriminators should be the same.

We note that similar ideas of employing multiple discriminators has already been raised by \cite{LaplaceGANs, CRN, pix2pixHD}. They resize the real/fake images to multiple scales and apply discriminator subnetworks with the same architecture to sense different levels of structures of the real/fake images.
In comparison, we fix the size of the real/fake images, and apply discriminators of different depths to achieve multiple sizes of receptive fields. It is more stable and computationally efficient to share the weights of the shallow convolutional layers. Comparison experiments with different numbers of discriminator subnetworks and previous multi-scale discriminators are described in Sec.\ref{subsec:diff_d}. 

\subsection{Loss Function}
\label{subsec:losses}

With the multi-scale discriminator $D$, which consists of subnetworks $\{D_i, i=0,1,\cdots, N_D\}$, the adversarial loss is written as 
\begin{equation}
\label{eqn:new_loss_adv}
\begin{aligned}
\mathcal{L}_{adv}(G;D)&=\frac{1}{N_D}\sum_{i=0}^{N_D}E_{(\bm{x},\bm{y})\sim p_{data}(\bm{x},\bm{y})}\big[\log D_i(\bm{x},\bm{y})\big] \\
& + E_{\bm{x}\sim p_{data}(\bm{x})}\Big[\log \Big(1-D_i \big(\bm{x},G(\bm{x})\big)\Big)\Big].
\end{aligned}
\end{equation}

Similar to the pix2pix model, we also adopt an $L1$ loss to encourage the generated image $G(\bm{x})$ from a line map $\bm{x}$ to be close to its ground truth image $\bm{y}$.  
The $L1$ loss is given by
\begin{equation}
\label{eqn:loss_l1}
\mathcal{L}_{L1}(G)=\mathbb{E}_{(\bm{x},\bm{y})\sim p_{data}(\bm{x},\bm{y})}\big[\|\bm{y}-G(\bm{x})\|_1\big].
\end{equation}

In order to achieve better perceptual quality in generated face images, we add a feature matching loss $\mathcal{L}_{fm}$. 
The feature matching loss~\cite{pix2pixHD} is a variant of perceptual loss~\cite{PerctualMetrics, StyleTransfer, PerceptualLosses}, which aims to minimize errors between generated images and corresponding ground-truth images in the feature space. 
Different from the previous techniques that employ perceptual loss through a pretained VGG model, the feature matching loss uses the feature maps produced by the discriminator in our CSAGAN. 
This is because the VGG models used in previous methods are always trained with the ImageNet dataset and have domain gaps with face images.
Our discriminator is trained specially for face images; therefore, it is more suitable for extracting features that present perceptual information of faces. 
Specifically, let $D^q_i(\cdot)$ be the output of the $q$-th layer of the $i$-th discriminator subnetwork with $n_i^q$ elements, and the feature matching loss is given:
\begin{equation}
\label{eqn:feature_matching_loss}
\mathcal{L}_{fm}(G)=\frac{1}{N_DN_Q}\mathbb{E}_{(\bm{x},\bm{y})\sim p_{data}(\bm{x},\bm{y})}\sum_{i=0}^{N_D}\sum_{q\in Q} \frac{1}{n_i^q} \|D^q_i(G(\bm{x}))-D^q_i(\bm{y})\|_1 ,
\end{equation}
where $Q$ is the set of selected layers of discriminators, and $N_Q$ is the number of selected layers. 
We select the last three convolutional layers from each discriminator subnetwork in our experiments.

By combining the multi-scale discriminator and the feature matching loss, the full objective to train our CSAGAN is
\begin{equation}
	\label{eqn:new_minmax_game}
	\min_G \max_{D} \mathcal{L}_{adv}(G;D)+\lambda \mathcal{L}_{L1}(G) +\mu \mathcal{L}_{fm}(G).
\end{equation}
where $\lambda$ and $\mu$ are the weights to balance the three losses. We set $\lambda=100.0$, and $\mu=1.0$ in our experiments.

\subsection{Training Techniques}
Training GANs is non-trivial because it is hard for generator and discriminator networks to find equilibria in the adversarial minmax game. We apply several techniques to stabilize our training. 

\paragraph{Two-Timescale Update Rule (TTUR)} Previous works~\cite{FID, SAGANs} suggest that separate learning rates for the generator and the discriminator are able to compensate for the problem of slow learning in the discriminator. 
TTUR is applied in our training process and shown to be effective. 

\paragraph{Spectral Normalization} Spectral normalization~\cite{SN} is a recently proposed normalization technique, which restricts the spectral norm in each layer of the discriminator to constrain its Lipschitz constant. Spectral normalization is computationally efficient and requires no extra hyper-parameter. Furthermore, spectral normalization is beneficial for generator training because it prevents unusual gradients~\cite{SAGANs}. 
We apply spectral normalization for both our generator and discriminators.

\paragraph{Multi-Stage Training} In order to stabilize training,, we divide the training process into three stages. In the first stage, we train the model without CSAM. 
Then, we add CSAM and train the CSAM while fixing the weights of the other layers in the second stage. 
Finally, we fine-tune the entire CSAGAN model together.

	\section{Experiment}
\label{sec:experiment}
We apply the proposed CSGAN framework to generate realistic photos from sparse line drawings of faces. We conducted a series of experiments to demonstrate the effectiveness of our method for preserving facial structures and generating fine details. Comparisons with other state-of-the-art methods also show the superiority of the proposed CSGAN. 

\paragraph{Dataset}
To train our network, we use the CelebA-HD dataset~\cite{PGGAN}, which contains 30K high-resolution celebrity images. We randomly select 24K images for training and 6K for testing. All the images are resized to $256 \times 256$ in our experiments. 
To generate pairs of line drawings and face photos for supervised training, we adopt a pipeline similar to pix2pix.
Specifically, edges are first extracted using a deep edge detector named Holistically-nested Edge Detector (HED)~\cite{HED}. 
In a generated edge map, each pixel has a value ${p_{HED}}$ indicating the probability of it being an edge. 
Several post-processing steps, including thinning, short edge removal, and erosion, are conducted to obtain simpler and clearer line maps with fewer edge fragments.

\paragraph{Training Parameters.}
We use the Adam~\cite{Adam} optimizer with momentum parameters $\beta_1=0.5, \beta_2=0.999$. We update one step for either $G$ or $D$ alternatively, and batch size is set to 8. 
Either the first or the second training stage lasts 100 epochs with an initial learning rate $lr_G=0.0001$ for the generator and $lr_D=0.0004$ for the discriminator, while the third stage lasts 50 epochs with initial learning rates $lr_G=0.00001$ and $lr_D=0.00004$. 
The learning rates decay at the halfway point of each stage. 
The entire training process takes about seven days on eight GeForce GTX 1080Ti GPUs.
\subsection{Evaluation Metrics}
\label{subsec:evaluation}
The evaluation of generative models is an open and complicated task.
A model with good performance with respect to one criterion does not necessarily imply good performance with respect to another criterion~\cite{evaluation, GANs_equal}. 
Traditional metrics, such as pixel-wise mean-squared error, do not present the joint statistics of the synthesized samples, and therefore are not able to evaluate the performance of a conditional generated model. 

Since the goal of our lines-to-image translation is to generate face images that are visually plausible, we compare the results of different models with perceptual user studies, which are commonly used for evaluating GAN models~\cite{LaplaceGANs, SRGANs, Improved_Techniques, CRN, pix2pixHD}. 
Following a similar procedure as described in \cite{CRN}, we conduct two kinds of experiments: an unlimited time user study and a limited time user study. 
In addition, we use three popular quantitative evaluation metrics, the inception score (IS)~\cite{Improved_Techniques}, Fr\'echet Inception Distance (FID)~\cite{FID}, and Kernel Inception Distance (KID)~\cite{KID}, which are proved to be consistent with human evaluation in assessing the realism of images. More details are explained below.

\paragraph{User Study with Unlimited Time.}
In every trial of the user study with unlimited time, we randomly select a conditional line map from the testing dataset and generate two synthesized images using two approaches. 
The two synthesized images are displayed randomly on the left and right side of the conditional line map.
The user has unlimited time to pick the one that "is more realistic and matches the conditional image better". 
No feedback is provided after each trial to avoid disturbing the user's perceptual judgment and preference.

\paragraph{User Study with Limited Time.}
In the study with limited time, we evaluate how quickly users perceive the differences between images. 
For each line map, we obtain four corresponding face images (one ground truth image and three synthesized images generated by our method, pix2pix~\cite{pix2pix}, and  SketchyGAN~\cite{SketchyGANs}, respectively).
In each comparison, we randomly select two images from these four corresponding face images and show the two selected images with the conditional line map.
Similarly, the two images are displayed to the user on the left and right sides of the line map randomly.
Within a duration randomly selected from a set of $\{1/8, 1/4, 1/2, 1, 2, 4, 8\}$ seconds, the user is asked to pick the one that "is more realistic and matches the conditional image better."  
We compute the percentage of the results generated by different methods that are preferred with different time durations. 
\paragraph{Inception Score (IS)} 
IS~\cite{Improved_Techniques} computes the KL divergence between the conditional class distribution and the marginal class distribution. Although it has been pointed out that IS has serious limitations because it focuses more on the recognizability of generated images rather than the realism of details or intra-class diversity \cite{evaluation}, it is still widely used to compare the quality of generated images.
\paragraph{Fr\'echet Inception Distance (FID)}
FID~\cite{FID} is a recently proposed and widely used evaluation metric for generative models.
It is shown to be consistent with human perceptual evaluation in assessing the realism of generated samples. 
It employs an Inception network to extract features and calculates the Wasserstein-2 distance between features of generated images and real images. 
Lower FID values indicate that the synthetic distribution is closer to the real distribution. 
\paragraph{Kernel Inception Distance (KID)} 
Similar to the FID, KID~\cite{KID} measures the difference between two sets of samples by calculating the squared maximum mean discrepancy between Inception representations. Moreover, unlike the FID, which is reported to be empirically biased, KID has an unbiased estimator with a cubic kernel~\cite{KID}, and it matches human perception more consistently.
\subsection{Comparisons with Previous Methods}
We compare the proposed model with two state-of-the-art lines-to-photo translation methods, pix2pix~\cite{pix2pix} and SketchyGAN~\cite{SketchyGANs}. 
We train the pix2pix model on our edge-face dataset using the default settings described in \cite{pix2pix}. 
SketchyGAN is originally designed for multi-class sketch-to-image generation.
We remove its classification branch and the loss term for classification, and train the pruned network with our line-face dataset. 
Firstly, the unlimited time user study was conducted to evaluate the perceptual quality of generated images. 
50 users participated in these experiments, and each user was tested with about 250 trials. 
The results are reported in Table \ref{table:unlimited_time}. 
These results show that given unlimited time, users are able to discover the visual differences between the generated face images using our model and previous ones. 
Compared with pix2pix and SketchyGAN, our results are significantly preferable according to the test users.

\begin{figure}
	\includegraphics[width=0.5\textwidth]{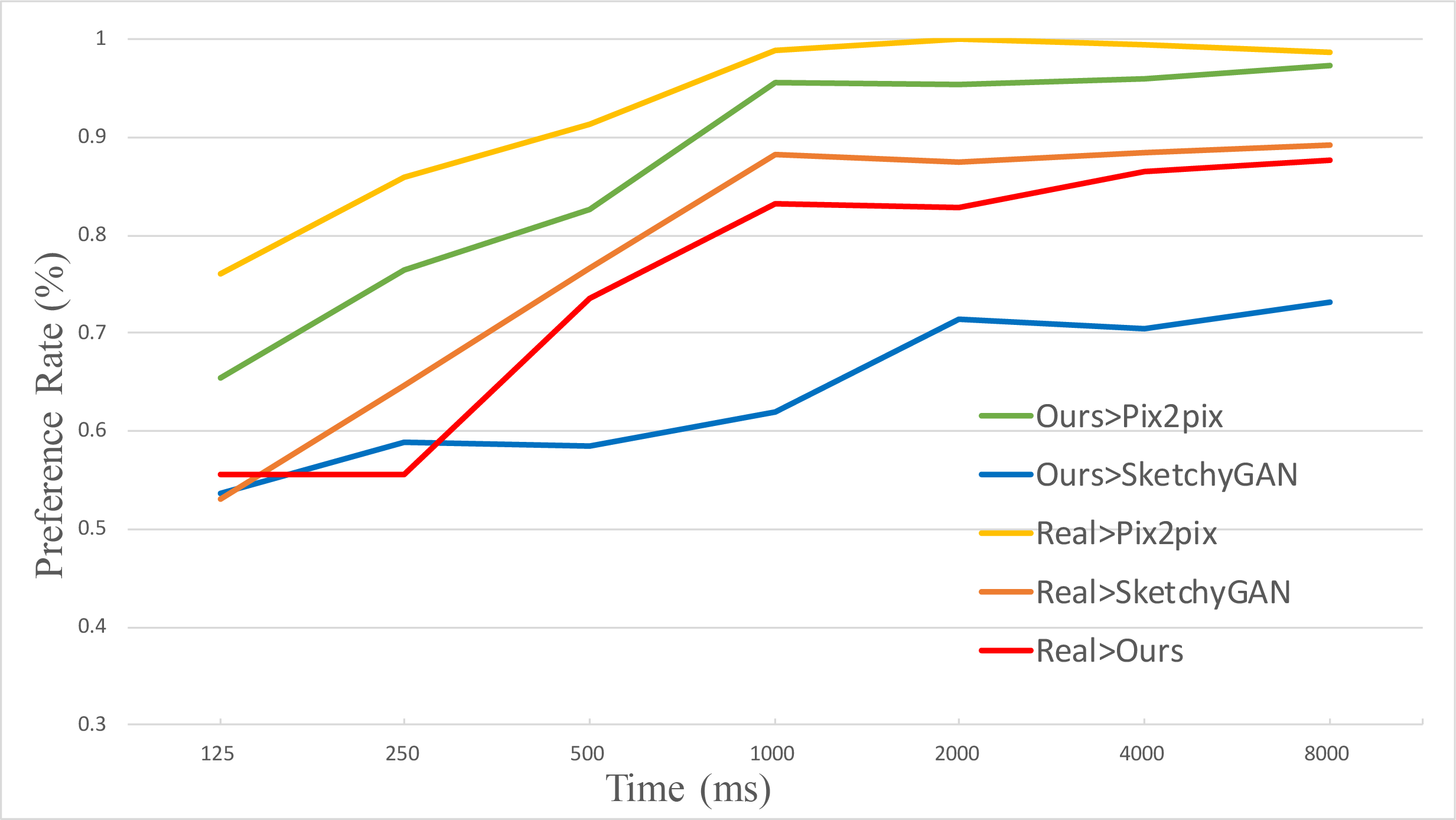}
	\caption{The results of limited time user studies. Each line indicates the user preference rate of one method over another. Users observe more differences between these methods as the display time lengthens.. }
	\label{fig:limited_time}
\end{figure}

Secondly, the limited time user studies were conducted to evaluate how quickly users can perceive the differences between images generated by different methods. 
Figure \ref{fig:limited_time} shows the results. 
When images are shown for a very short time ($1/8$ seconds), the users are not able to sense the differences among different methods and the ground truth.As the time increases, more differences are perceived by users, and more users prefer the results generated by our CSAGAN.

Thirdly, Table~\ref{tab:evaluation_metrics} lists the quantitative comparisons of our method and others. 
As we can see, our full model surpasses the pix2pix model and SketchGAN by a large margin with regard to the mean values of IS/FID/KID, demonstrating our model's capability to generate more realistic face photos.

\begin{figure}
	\includegraphics[width=\linewidth]{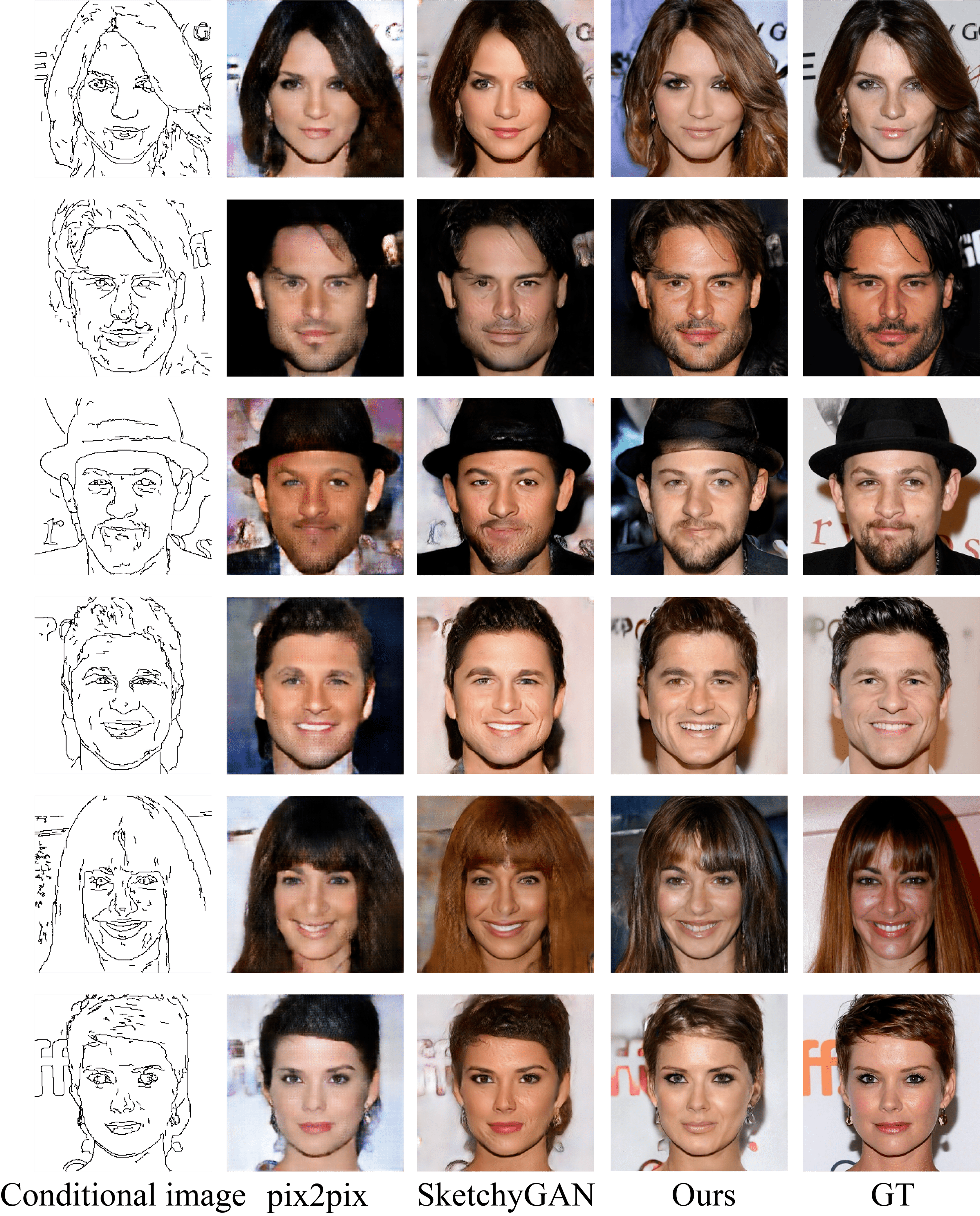}
	\caption{The above face images are generated from edge maps using three methods: pix2pix~\cite{pix2pix}, SketchyGAN~\cite{SketchyGANs} (without a classification network), and our proposed method. Ground truth (GT) images that we use to obtain the edge maps are shown in the right-most column. The results generated by our model contain more details, especially in the areas with hairs, whiskers, and highlighted regions. Also, our results appear more realistic with regard to the illumination of faces.}
	\label{fig:results}
\end{figure}

Finally, Figure~\ref{fig:results} shows a group of synthesized face images using the proposed model and previous methods (better viewed in color). We observe that the results of our model contain more details, especially in the areas with hairs, whiskers, and highlighted regions, while the results of the previous models are over-smoothed and lack realistic details.
Moreover, our results appear more realistic with 
regard to the illumination of faces.
\begin{table}[h]
	\centering	
	\caption{Results of user study with unlimited time.}
	\begin{tabular}{|l|c|c|}\hline
		& Pix2pix \cite{pix2pix} vs Ours & SketchyGAN\cite{SketchyGANs} vs Ours \\\hline
		Preference&$6.9\% / 93.1\%$&$24.0\% / 76.0\%$\\\hline
	\end{tabular}
	\label{table:unlimited_time}
\end{table}

\begin{table}[h]
	\centering	
	\caption{Quantitative comparison. (Larger IS and lower FID/KID represent better results.)}
	\begin{tabular}{|l|c|c|c|}\hline
		Models & IS & FID & KID \\\hline
		Pix2pix \cite{pix2pix} & $2.55\pm0.20 $ & $605.97\pm13.95 $ & $3.05\pm0.08 $ \\
		SketchyGAN \cite{SketchyGANs} &$2.75\pm0.17 $ & $479.09\pm15.24 $ & $2.11\pm0.09 $ \\\hline
		Ours w/o CSAM & $2.65\pm0.08 $ & $426.78\pm17.23 $ &$1.62\pm0.04 $ \\
		Ours w/ $D_p$ & $2.73\pm0.09 $ & $413.26\pm13.92 $  & $1.57\pm0.07 $ \\\hline
		Ours w/ $D_{pix2pixHD}$ & $2.70\pm0.08 $ & $398.78\pm14.60 $  & $1.40\pm0.05 $ \\
		Ours, $N_D=2$  & ${2.71\pm0.13 }$   & ${332.00\pm9.26 }$  & ${0.98\pm0.04 }$  \\
		Ours, $N_D=3$ & $\mathbf{2.78\pm0.09 }$   & ${269.96\pm8.18 }$  & ${0.63\pm0.04 }$  \\
		Ours, $N_D=4$ & ${2.76\pm0.10 }$   & $\mathbf{269.59\pm8.89 }$  & $\mathbf{0.62\pm0.05 }$  \\
		\hline
	\end{tabular}
	\label{tab:evaluation_metrics}
\end{table}

\subsection{Ablation Study}
\label{subsec:ablation}

\begin{figure}
	\includegraphics[width=\linewidth]{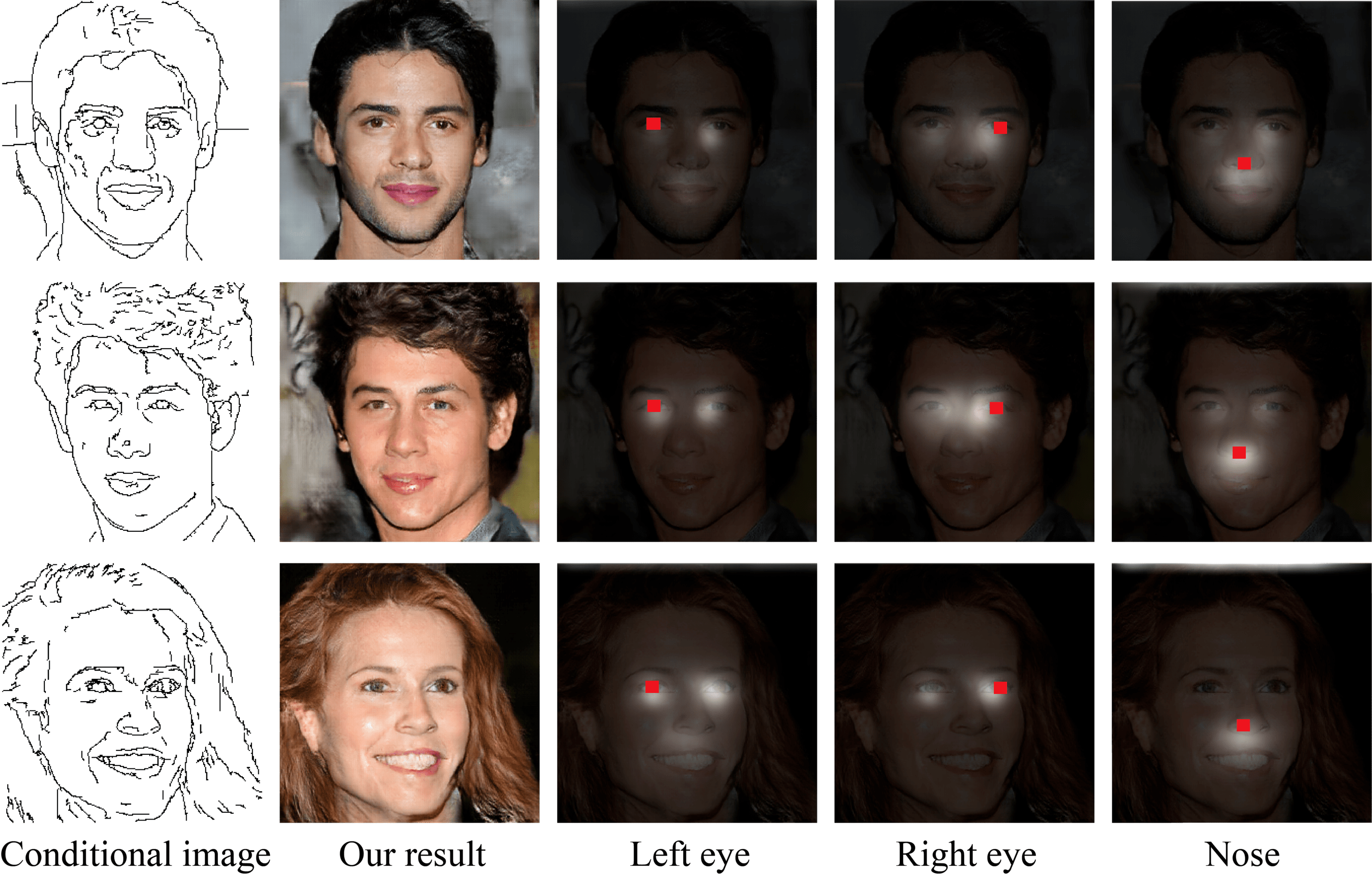}
	\caption{The attention maps are shown with the conditional edge maps and the generated images. Three locations of the nose and two eyes are marked in red while the attention maps with respect to these locations are shown. The larger values in the attention maps are brighter in the figure. We observe that long-range dependence between different parts of faces is captured by our CSAM.}
	\label{fig:attention}
\end{figure}

We examine the importance of every component within our model based on IS/FID/KID, shown in the third and fourth row of Table \ref{tab:evaluation_metrics}. The experiments were conducted by removing each specific part from the full model and then training the rest of th model without the absent part. Specifically, we remove 1) the proposed CSAM (ours w/o CSAM), and 2) the multi-scale discriminator, using only the patch discriminator $D_p$ (ours w/ $D_p$). As we can see, the performance of our model without CSAM drops dramatically compared to the full model, indicating the critical importance of CSAM in our model. The performance of our model also benefits from the multi-scale discriminator.
We also visualize the attention maps to demonstrate how pixels in different locations are related and dependent in the learned model.
Figure \ref{fig:attention} shows a group of examples of attention maps. 
Three locations  (i.e., the nose and the two eyes) are marked in red, and the attention maps with respect to these locations are shown, respectively. 
The larger values in the attention maps are brighter in the figure. We observe that the long-range dependence is captured by the CSAM. For example, to generate the pixels in one eye, the regions of both eyes are assigned high attentions. In another words, the information for generating a specific pixel comes from not only its local area but also related regions far away from this pixel.

\subsection{Different Levels of Details in Line  Maps.}
\label{subsec:gamma}

\begin{figure}
	\includegraphics[width=\linewidth]{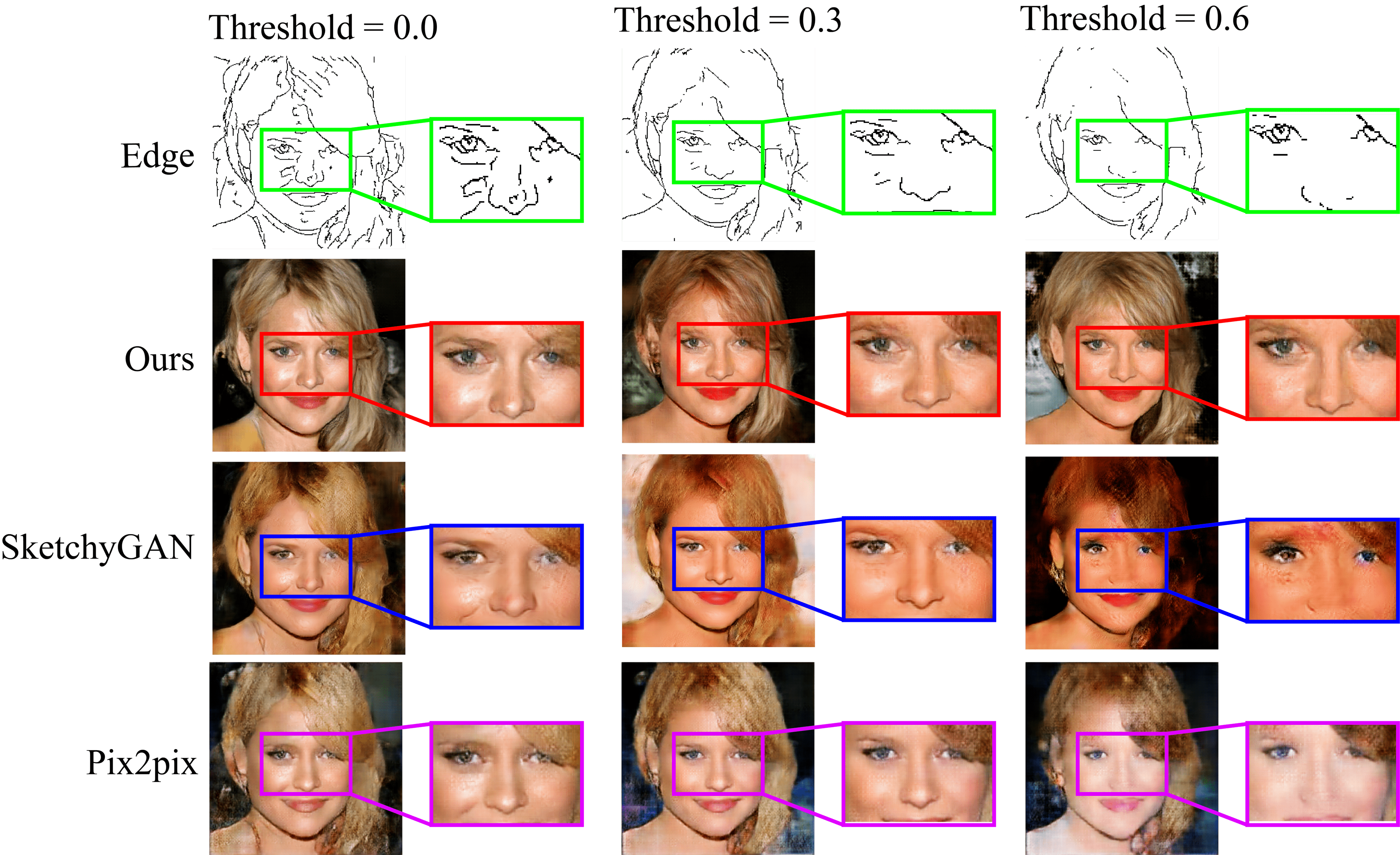}
	\caption{ The face images generated from line maps with different levels of details. Our proposed model is able to generate realistic face images with complete structures and fine textures (the second row). Whereas, the two previous models~\cite{pix2pix, SketchyGANs} fail to generate the nose and the left eye when $\tau$ is set to 0.3 and 0.6. The area around the nose is zoomed in, and the ground truth is displayed on the right. }
	\label{fig:gamma}
\end{figure}

To evaluate the robustness of our CSAGAN, we use line maps with different levels of details to produce face images.
As discussed in the dataset construction, we produce edge maps based on $p_{HED}$ with several post-processing steps. 
The lines in each edge map are generated by keeping edge pixels that are $p_{HED}>\tau$. A larger $\tau$ value causes less detail in the edge maps.
By setting different values for $\tau$, we generate edge maps with different levels of details, as shown in Figure~\ref{fig:gamma}.
The proposed model is robust enough to generate face images with the whole structure when the inputs are line maps with different levels of details.
In comparison, the two previous models fail to generate some parts of the face (i.e., the nose) when detail edges are missing in the line maps with larger $\tau$ value (0.3 and 0.6 in this case).

\subsection{Comparison of Different Multi-scale Discriminators}
\label{subsec:diff_d}
We compare our multi-scale discriminator with its variants and the one from previous work~\cite{pix2pixHD}. More specifically, if we let $N_D$ be the number of discriminator subnetworks, we train the models with the same generator and discriminator as $N_D=\{2,3,4\}$ subnetworks, denoted as ours, $N_D=\{2,3,4\}$. Each subnetwork shares weights with others in the first few layers, while the depths are different. Therefore, the receptive fields of the last layers in the subnetworks are different, and the subnetworks distinguish generated samples from real ones in different scales. Quantitative comparison results based on IS/FID/KID are shown in the last three rows of Table~\ref{tab:evaluation_metrics}. 

Also, we compare our multi-scale discriminator with the one in \cite{pix2pixHD}. Specifically, we use our generator and switch our discriminator to the one from \cite{pix2pixHD} ($N_D=3$) and train this model with our three-stage training process, which is denoted as ours w/ $D_{pix2pixHD}$. Results shown in Table~\ref{tab:evaluation_metrics} indicate that our multi-scale discriminator (ours, $N_D=3$) exceeds its counterpart in the measure of IS/FID/KID and shows its advantages on quantitative evaluation. 
	\section{Conclusion}
\label{sec:conclusion}
In this work, we propose a conditional self-attention GAN (CSAGAN) to synthesize photo-realistic face images from sparse lines. By introducing the self-attention mechanism and a multi-scale discriminator into conditional GANs, our method is able to capture long-range dependence across different regions and global structures in face images. Comprehensive experiments illustrate the effectiveness of the proposed method via two perceptual studies and three quantitative metrics. 
Our framework shows its promising capability to generate high-quality face images by synthesizing complete facial structures as well as fine details, even when some parts of the input line map are missing. 

	\begin{acks}
This work was supported by the National Key Research $\&$ Development Plan of China under Grant 2016YFB1001402, the National Natural Science Foundation of China (NSFC) under Grants 61632006, 61622211, and 61620106009, as well as the Fundamental Re-search Funds for the Central Universities under Grants WK3490000003 and WK2100100030.
\end{acks}

	\bibliographystyle{ACM-Reference-Format}
	\balance 
	\bibliography{SCAGANs_NEW}
	
\end{document}